\documentclass[letterpaper, 10 pt, conference]{ieeeconf}  

\IEEEoverridecommandlockouts                              

\usepackage{comment}
\usepackage{amsmath} 
\usepackage{amssymb} 
\usepackage{graphicx}
\usepackage{balance}
\usepackage{color}
\usepackage{float} 
\usepackage[colorinlistoftodos]{todonotes} 
\usepackage{subcaption}
\usepackage{tabularx} 
\usepackage{hyperref}
\usepackage{gensymb}
\usepackage{ifthen} 

\title{\LARGE \bf
Inclined Surface Locomotion Strategies for Spherical Tensegrity Robots}

\author{Lee-Huang Chen$^{1}$, Brian Cera$^{1}$, Edward L. Zhu$^{1}$, Riley Edmunds$^{1}$, Franklin Rice$^{1}$,\\
Antonia Bronars$^{1}$, Ellande Tang$^{1}$, Saunon R. Malekshahi$^{1}$, Osvaldo Romero$^{1}$,\\
Adrian K. Agogino$^{2}$ and Alice M. Agogino$^{1}$
\thanks{*Research supported by NASA Early Stage Innovation grant NNX15AD74G}
\thanks{$^{1}$Authors are with the University of California, Berkeley, Mechanical Engineering Department,
        CA, 94720, USA 
        Contact email: {\tt\small leehuanc@berkeley.edu}}%
\thanks{$^{2}$Adrian K. Agogino is with NASA Ames, Moffett Field, CA 94035 USA
        {\tt\small adrian.k.agogino@nasa.gov}}%
}

\begin{document}

\maketitle
\thispagestyle{empty}
\pagestyle{empty}


\begin{abstract}

This paper presents a new teleoperated spherical tensegrity robot capable of performing locomotion on steep inclined surfaces. With a novel control scheme centered around the simultaneous actuation of multiple cables, the robot demonstrates robust climbing on inclined surfaces in hardware experiments and speeds significantly faster than previous spherical tensegrity models. This robot is an improvement over other iterations in the TT-series and the first tensegrity to achieve reliable locomotion on inclined surfaces of up to 24\degree. We analyze locomotion in simulation and hardware under single and multi-cable actuation, and introduce two novel multi-cable actuation policies, suited for steep incline climbing and speed, respectively. We propose compelling justifications for the increased dynamic ability of the robot and motivate development of optimization algorithms able to take advantage of the robot's increased control authority.

\end{abstract}

\section{Introduction}

UC Berkeley and the NASA Ames Research Center are developing a new concept for space exploration robots based on tensegrity structures. A tensegrity structure consists of rods suspended in a network of cables, where the rods and cables experience only compression and tension, respectively, while in equilibrium. Because there are no bending moments, tensegrity systems are inherently resistant to failure \cite{skelton2001}. Additionally, the structures are naturally compliant, exhibiting the ability to distribute external forces throughout the tension network. This mechanical property provides shock protection from impact and makes the structure a robust robotic platform for mobility in an unpredictable environment. Thus, tensegrity robots are a promising candidate for exploration tasks, especially in the realm of space exploration, because the properties of tensegrity systems allow these robots to fulfill both lander and rover functionality during a mission.

Analysis of tensegrity robotic locomotion on inclined terrain is critical in informing path-planning and trajectory tracking decisions in mission settings.  Despite the crucial role of uphill climbing in planetary exploration, the \texorpdfstring{TT-4\textsubscript{mini}}{} robot is the first untethered spherical tensegrity robot to achieve reliable inclined surface climbing. 

\begin{figure}[t]
\center
\includegraphics[width=\columnwidth]{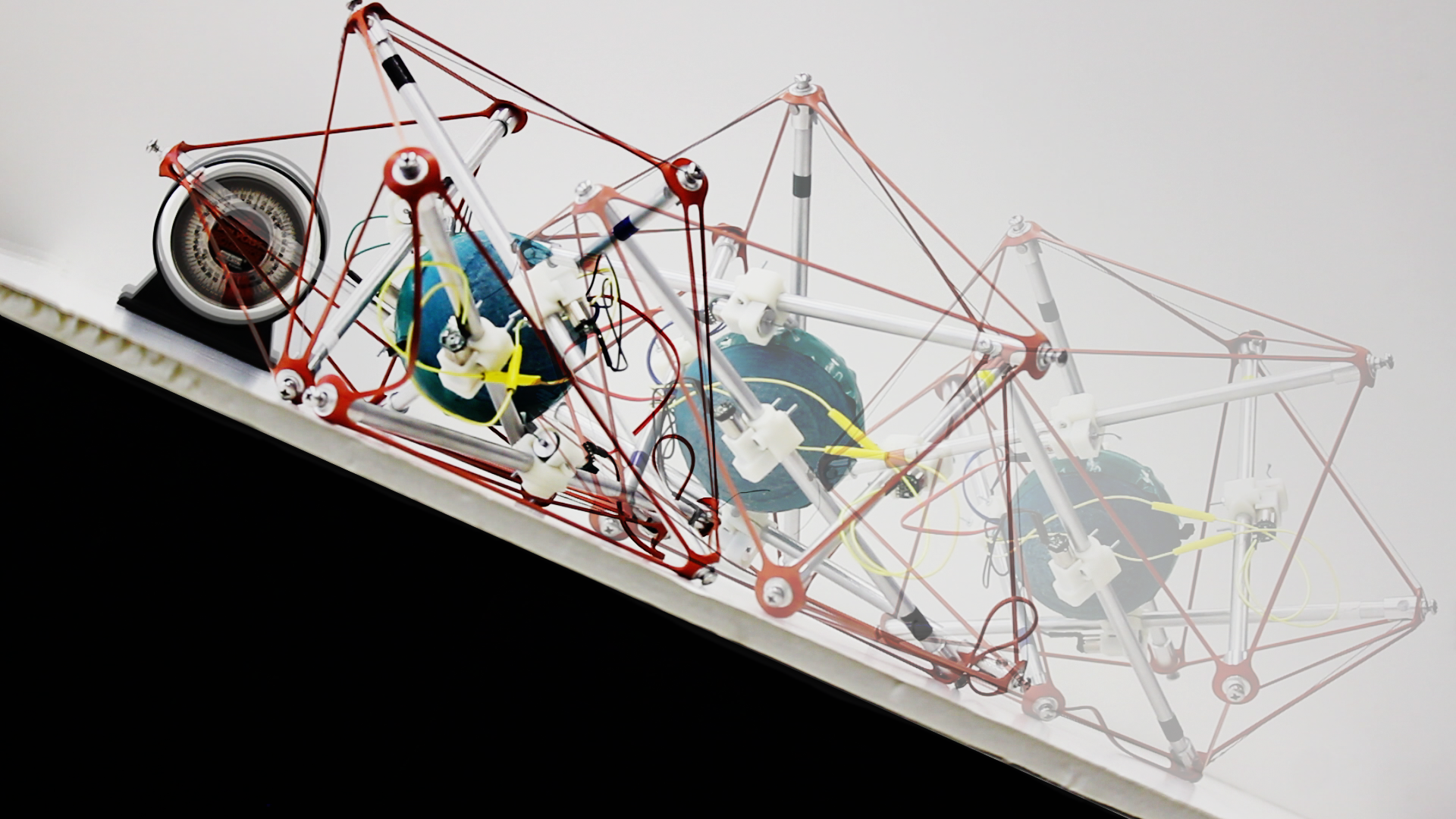}
  \caption{The \texorpdfstring{TT-4\textsubscript{mini}}{} prototype performing punctuated uphill rolling on an inclined surface of 24\degree.  The photo shows three steps by the robot.}
  \label{fig:TT4mini_rolling}
  \vspace{-0.5cm}
\end{figure}

The \texorpdfstring{TT-4\textsubscript{mini}}{} robot was rapidly constructed using a novel modular elastic lattice tensegrity prototyping platform \cite{elasticLattice}, which allows for rapid hardware iterations and experiments. This work presents the simulation results of inclined uphill locomotion for a six-bar spherical tensegrity robot as well as the prototyping and hardware experiments performed to validate these results. We show that the \texorpdfstring{TT-4\textsubscript{mini}}{} robot can achieve robust locomotion on surface inclines up to 24\degree~using a two-cable actuation scheme in hardware, as shown in Fig.~\ref{fig:TT4mini_rolling}.

In this paper, we first describe the topology and design of the \texorpdfstring{TT-4\textsubscript{mini}}{}, which uses a novel rapid tensegrity prototyping method. Next, we analyze incline locomotion performance in simulation under a single-cable actuation policy. This policy is tested on hardware to establish a performance benchmark against which two-cable actuation policies can be evaluated. Two variants of multi-cable policies are found in simulation, one suited for steep inclines and the other suited for speed. We demonstrate significant performance improvements in both tasks over the single-cable benchmark and discuss the primary factors that lead to improved performance. Finally, we motivate further research in extending this work to develop more efficient multi-cable locomotion policies by leveraging learning algorithms.

\section{Prior Research}

Tensegrity robots have become a recent subject of interest due to their applications in space exploration \cite{Caluwaerts2014}. The natural compliance and reduced failure modes of tensegrity structures have motivated the development of multiple tensegrity robot forms \cite{skelton2001}. Some examples include spherical robots designed for locomotion on rugged terrain \cite{Koizumi2012b,Kim2014,Kim2015}, snake-like robots that crawl along the ground \cite{Mirletz2015a}, and assistive elements in walking quadrupedal robots \cite{Sabelhaus2015,Hustig-Schultz2016,Sabelhaus2017a}.

Tensegrity locomotion schemes have been studied in both the context of single-cable actuation \cite{planartensegrity}, and (rarely) in the context of multi-cable actuation \cite{dynamiccomplexity}. However, much of this exploration into tensegrity multi-cable actuation policies has been in the context of vibrational, rather than rolling motion. 

While there exists extensive prior work in incline robotic locomotion, this literature does not directly address tensegrities. For example, Stanford's spacecraft/rover hybrid robot has demonstrated through simulation and hardware tests the potential for uphill locomotion. Rather than a tensegrity mechanism, however, Stanford's hybrid robot uses a flywheel-based hopping mobility mechanism designed for traversing small micro-gravity bodies \cite{hedgehog}.

Movement on rough or uphill terrain is a frequent occurrence in space exploration, and has proven to be a necessary challenge for traditional wheeled rovers. For instance, Opportunity has ascended, with much difficulty, a number of surfaces up to 32\degree~above horizontal \cite{JPLOpportunity}. On the other hand, NASA's SUPERball, which is also a 6-bar tensegrity robot, has demonstrated successful navigation of an 11.3\degree~(20\% grade) incline in simulation \cite{SuperBall}. However, as will be discussed later, the \texorpdfstring{TT-4\textsubscript{mini}}{} is the first tensegrity robot to successfully demonstrate significant inclined surface locomotion, not only in simulation, but also in hardware testing. 

\section{Six-bar Tensegrity Robot Using Modular Elastic Lattice Platform}

\begin{figure}[htb]
    \center
    \includegraphics[width=\columnwidth]{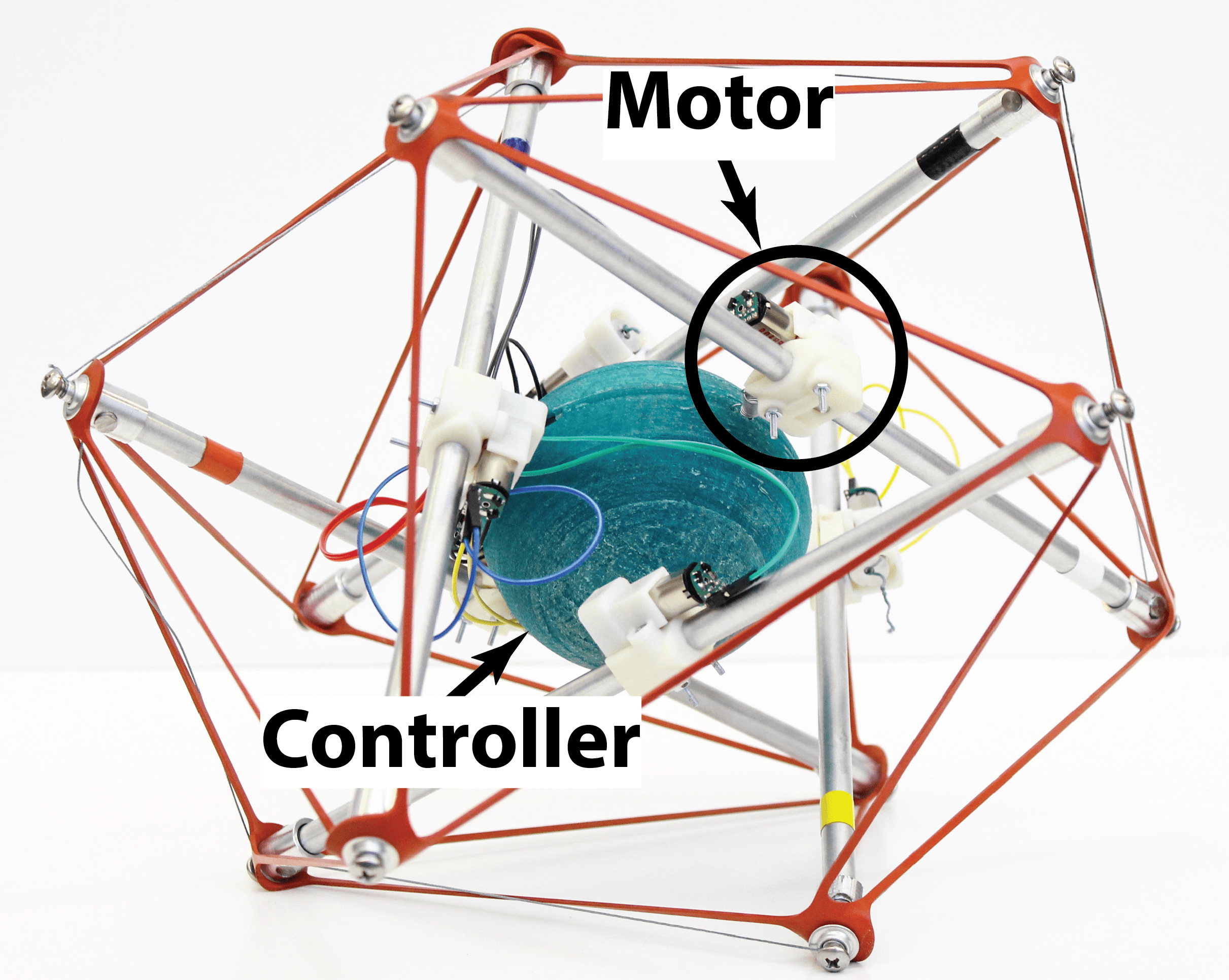}
    \caption{The \texorpdfstring{TT-4\textsubscript{mini}}{} robot in its neutral stance.  The tensegrity robot was assembled by adding six actuators and a central controller to the static modular elastic tensegrity structure. The \texorpdfstring{TT-4\textsubscript{mini}}{} achieves shape-shifting by contracting or releasing cables connected in parallel with the elastic lattice.}
  \label{fig:TT4mini}
  \vspace{-0.3cm}
\end{figure}

In order to greatly simplify and expedite the process of assembly we developed a modular elastic prototyping platform for tensegrity robots \cite{elasticLattice}. The \texorpdfstring{TT-4\textsubscript{mini}}{}, a six-bar spherical tensegrity robot, was the first tensegrity robot assembled using this new prototyping platform and can be rapidly assembled in less than an hour by a single person. To construct the robot, a regular icosahedron structure is first rapidly assembled using the modular elastic lattice platform and six aluminum rods of 25 cm each, creating the passive structure of the tensegrity robot. A total of six actuators and a central controller are then attached to the structure, resulting in a dynamic, underactuated tensegrity robot, as shown in Fig.~\ref{fig:TT4mini}.

\section{Single-Cable Actuated Climbing on Inclined Surfaces}

A spherical tensegrity robot can perform rudimentary punctuated rolling locomotion by contracting and releasing each of its cables in sequence, deforming its base and shifting its center of mass (CoM) forward of the front edge of its supporting base polygon. This contraction places the robot in a transient, unstable state, from which it naturally rolls onto the following stable base polygon. After the roll, the robot releases the contracted cable and returns to its neutral stance before initiating the next step in the sequence. In this paper, the neutral stance of the robot refers to the stance in which no cables are contracted and the only tension in the system is due to gravity.

While other robots have successfully achieved punctuated rolling on flat ground using this technique \cite{Chen2016,SuperBall}, we show that the \texorpdfstring{TT-4\textsubscript{mini}}{} is not only capable of the same, but can also do so on an inclined surface. This section summarizes the results for a single-cable actuation policy and sets the standards against which we evaluate the improved climbing capabilities achieved through two-cable actuation (Section V).

\subsection{Simulation and Analysis of Single-Cable Actuation Policies}

\begin{figure}[t]
    \center
    \includegraphics[width=0.9\columnwidth]{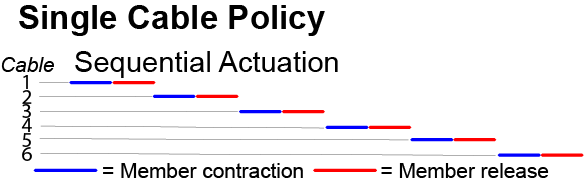}
    \caption{Visualization of the single-cable actuation policy. Each row corresponds to one cable, this policy can be repeated indefinitely.}
    \label{fig:policies_one}
    \vspace{-0.3cm}
\end{figure}

\begin{figure}[t]
    \center
    \includegraphics[width=\columnwidth]{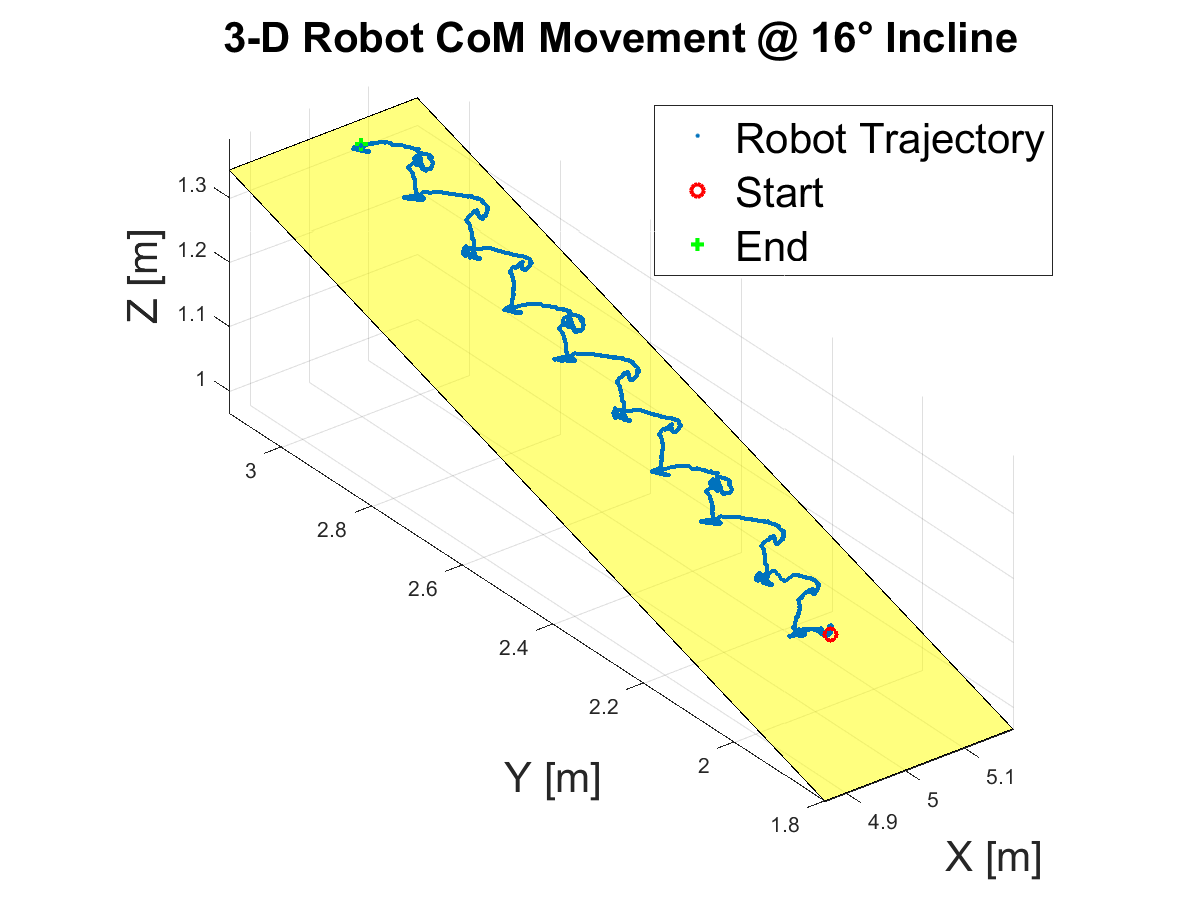}
    \caption{Simulation results of the \texorpdfstring{TT-4\textsubscript{mini}}{}'s payload CoM trajectory while climbing a 16\degree~incline using the single-cable actuation policy.}
  \label{fig:16deg3D}
  \vspace{-0.4cm}
\end{figure}

As there had been very little previous work on uphill climbing with spherical tensegrity robots, we first validated the actuation policy in simulation.  Using the NASA Tensegrity Robotics Toolkit (NTRT) simulation framework, we simulated the single-cable \texorpdfstring{TT-4\textsubscript{mini}}{} actuation scheme (Fig.~\ref{fig:policies_one}) for uphill climbing on surfaces of varying inclines.  Results showed that the robot could successfully climb an incline of 16\degree~in simulation using a single-cable actuation policy.  Simulation results at this incline are shown in Fig.~\ref{fig:16deg3D}. Beyond 16\degree, we found that the robot could no longer reliably perform locomotion, for the following two reasons: (1) The robot was unable to move the projected CoM sufficiently forward to initiate an uphill roll, and (2) Deformation of the base polygon shifted the CoM behind the back (downhill) edge of the polygon, initiating a downhill roll.

\begin{figure}[t]
    \center
    \includegraphics[width=0.8\columnwidth]{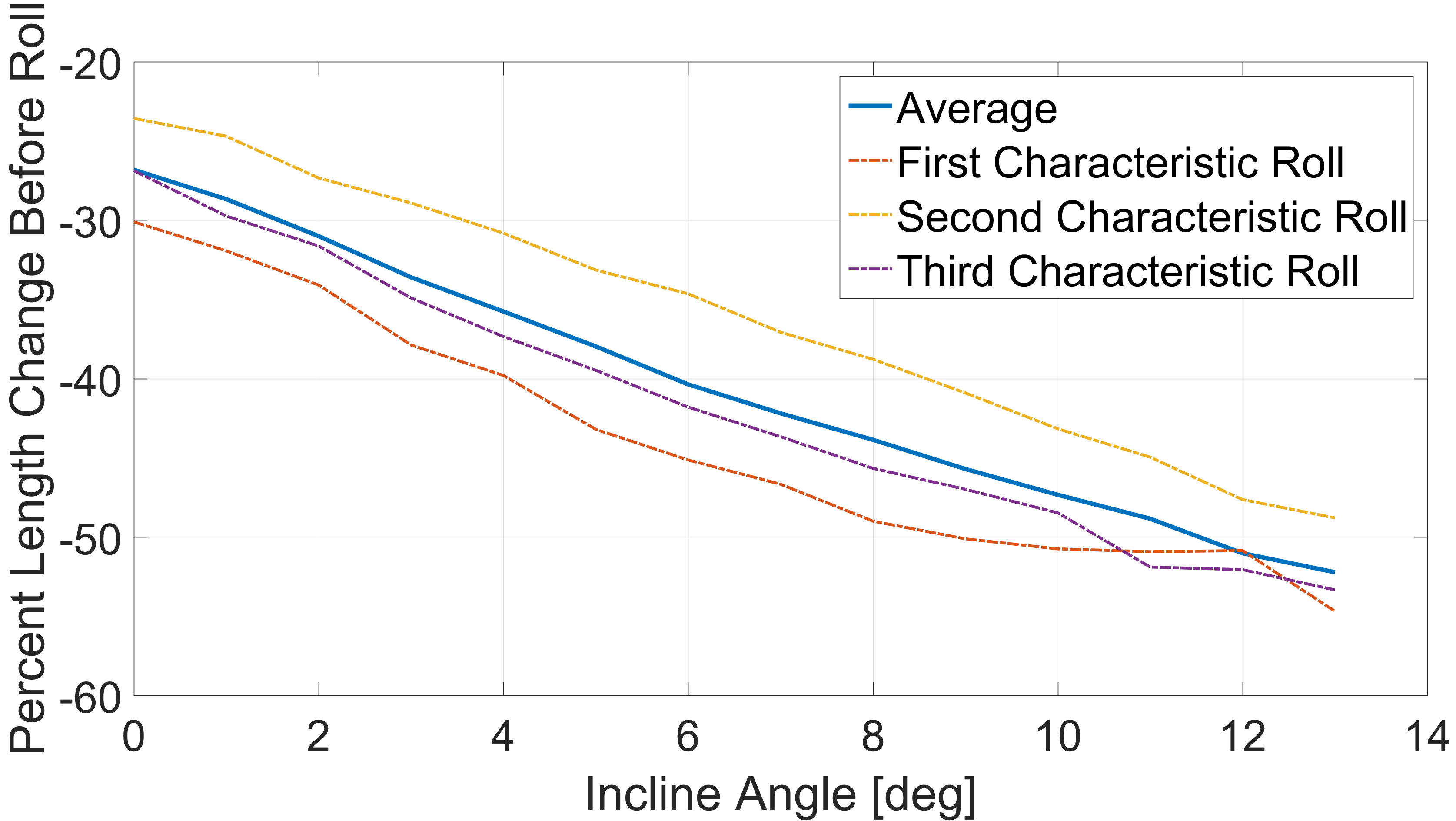}
    \caption{Required percent of cable retraction to initiate forward rolling motion with single-cable policy. Increasingly negative percentages signify greater cable retraction.}
  \label{fig:inclinecomparison}
  \vspace{-0.5cm}
\end{figure}

To analyze the limitations of single-cable actuation policies, we studied the relationship between actuation efficiency and incline angle using simulated sensor data.  At each angle of inclination, we recorded the cable actuation required to initiate rolling, as a fraction of initial cable length. As expected, we found that initiating tipping of the robot at greater angles of inclination requires greater cable contraction (Fig.~\ref{fig:inclinecomparison}). Interestingly, the extent to which the angle of inclination affects the required cable contraction is dependent on which particular cable is being actuated. Due to the inherent symmetry of the 6-bar spherical structure, the \texorpdfstring{TT-4\textsubscript{mini}}{}'s repeating six-step gait can be separated into two repeated three step sub-sequences, which arise from the uneven, yet symmetric, distribution of tensions in the springs suspending the central payload (in this case, the central controller). 

Our results imply that climbing steeper hills requires greater power consumption and more careful motion, motivating the development of more efficient actuation policies for uphill locomotion. This analysis highlights the mechanical limits of single-cable actuation policies, thus encouraging exploration of alternative actuation policies.

\subsection{Hardware Experiments of Single-Cable Actuation Policies}

In order to validate the results from software simulations, we constructed an adjustable testing platform which allowed for incremental adjustments of the surface incline angle. Using this setup, we considered as successful those trials in which the \texorpdfstring{TT-4\textsubscript{mini}}{} was able to reliably travel 91.4 cm (3 ft) along the inclined plane. We considered as failure those trials in which the \texorpdfstring{TT-4\textsubscript{mini}}{} failed to reach the 91.4 cm mark.  

We found that the robot was able to successfully perform uphill climbing up to 13\degree~in hardware with a single-cable actuation policy. Beyond 13\degree, relaxing a member after its successful contraction consistently shifted the CoM beyond the robot's backward tipping point, causing the structure to roll down the incline. The coefficient of static friction between the robot and the surface, measured for all 8 stable robot poses, ranged from 0.42 to 0.57 with a mean of 0.49. This corresponds to maximum inclines before slipping ranging from 23\degree to 29\degree, with a mean of 26\degree. We believe the reason for this range is due to the lack of material homogeneity at contact points between the robot and the ground, which consist of some combination of the rubber lattice and metal end-cap. In addition, as the distribution of weight on the end of the rods changes with the robot's orientation, it is likely that the frictional forces for each face are not uniform. 

Based on these results, we did not expect, nor did we observe any failure due to sliding in the single cable actuation tests. However, as will be discussed in later sections, this does become a limiting factor in the robot's performance at much steeper inclines. These results are consistent with failure modes observed in simulation.

As a baseline for comparison in later sections, the robot's average velocity was recorded when travelling 91.4 cm along a 10\degree~incline. Across 10 trials under these conditions, the \texorpdfstring{TT-4\textsubscript{mini}}{} achieved an overall average velocity of 1.96 cm/s. For reference the robot has a rod length of 25 cm. These results serve as the first demonstration of a tensegrity robot reliably climbing an inclined surface.

\section{Alternating and Simultaneous Two-Cable Actuated Climbing on Inclined Surfaces}

\begin{figure}[t]
    \center
    \includegraphics[width=0.9\columnwidth]{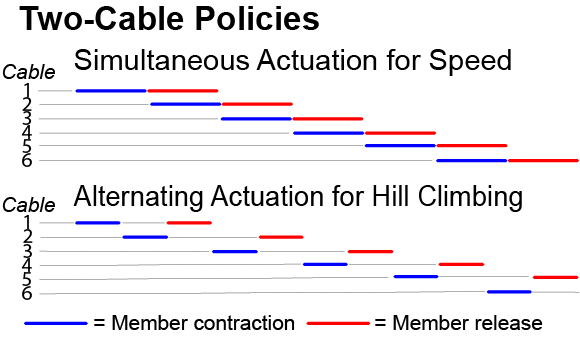}
    \caption{Visualizations of the two-cable actuation policies. Each row corresponds to one cable, and each policy can be repeated indefinitely.}
    \label{fig:policies}
    \vspace{-0.1cm}
\end{figure}

Having reached the limits of inclined locomotion for the single-cable actuation policy, the following actuation policies were explored:

\begin{itemize}
     \item \textbf{Simultaneous actuation policy}: Similar to single-cable actuation, except the next cable contracts as the current releases, allowing for more steps to be made in less time. See Fig.~\ref{fig:policies}.
    \item \textbf{Alternating actuation policy}: To preserve a low center of mass during uphill rolling, the next cable is fully contracted before the current is released. See Fig.~\ref{fig:policies}.
\end{itemize}

We found that multi-cable actuation policies allow the robot to climb steeper inclines and travel at significantly faster speeds than the single-cable actuation policy. The following sections present the performance results of two-cable actuation policies in simulation, and their validation through hardware experiments, summarized in Table \ref{table:1}.

\subsection{Simulation and Analysis of Alternating and Simultaneous Two-Cable Actuation Policies}

The two-cable actuation policies, as described above, were implemented and tested in NTRT as open-loop controllers using the same robot model and inclined surfaces as the aforementioned single-cable simulations. These simulations demonstrated vast improvements in incline locomotion stability as well as average speed, with the robot able to navigate inclines up to 26\degree~using alternating two-cable actuation and 24\degree~using simultaneous two-cable actuation.

\begin{table}[t]
\caption{Summary of Hardware Experiment Results}
\label{table:1}
\begin{center}
\begin{tabular}{ | c | c c c | }
\hline
Strategy & Avg. Speed@0\degree & Avg. Speed@10\degree & Max Incline \\
& [cm/s] & [cm/s] & [\degree] \\
\hline
Single & 3.19 & 1.96 & 13 \\
Simultaneous & 6.32 & 4.22 & 22 \\
Alternating & 3.02 & 2.12 & 24 \\
\hline
\end{tabular}
\end{center}
\vspace{-0.3cm}
\end{table}

\begin{figure}[t]
    \center
    \includegraphics[width=\columnwidth]{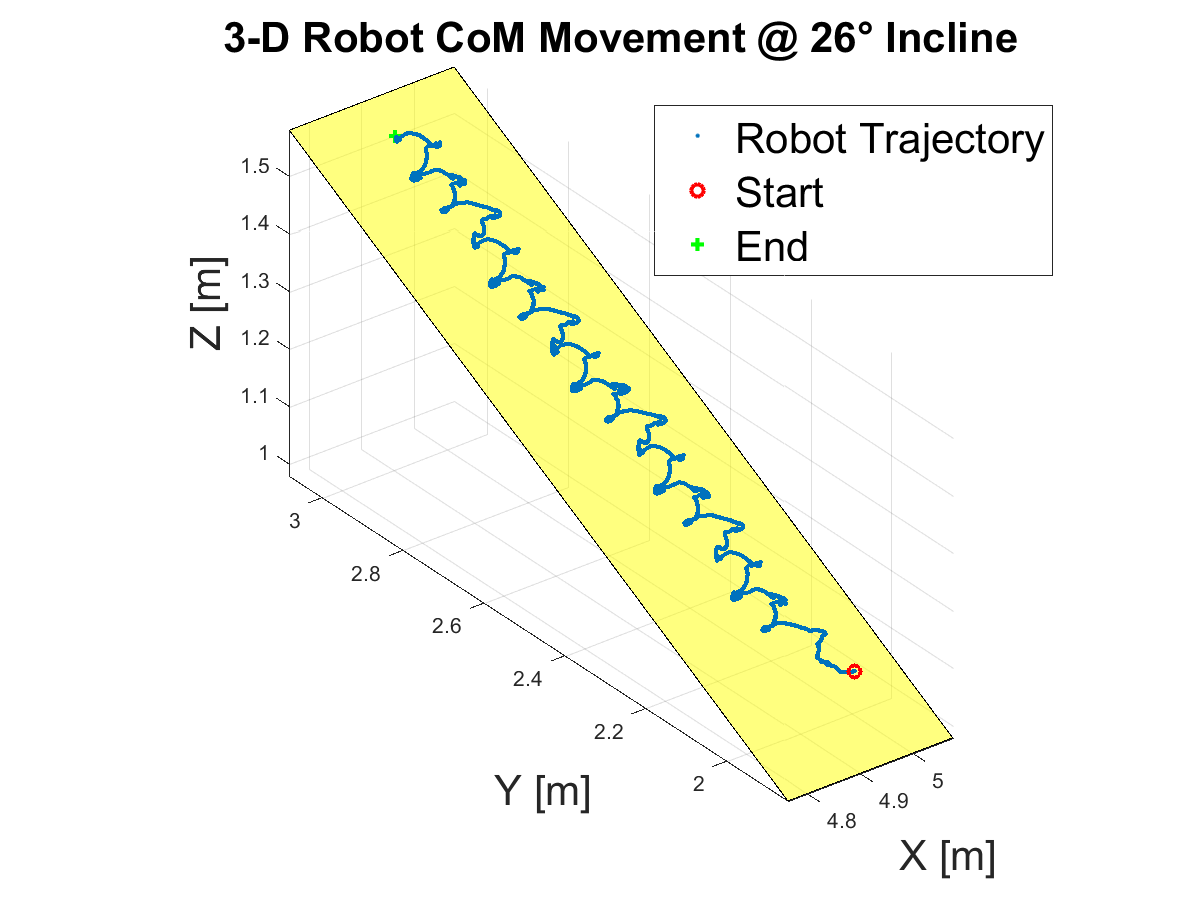}
    \caption{Simulation results of the \texorpdfstring{TT-4\textsubscript{mini}}{}'s payload CoM trajectory while climbing a 26\degree~incline using the alternating two-cable actuation policy.}
    \label{fig:26deg3D}
    \vspace{-0.3cm}
\end{figure}

\begin{figure}[h]
    \center
    \includegraphics[width=\columnwidth]{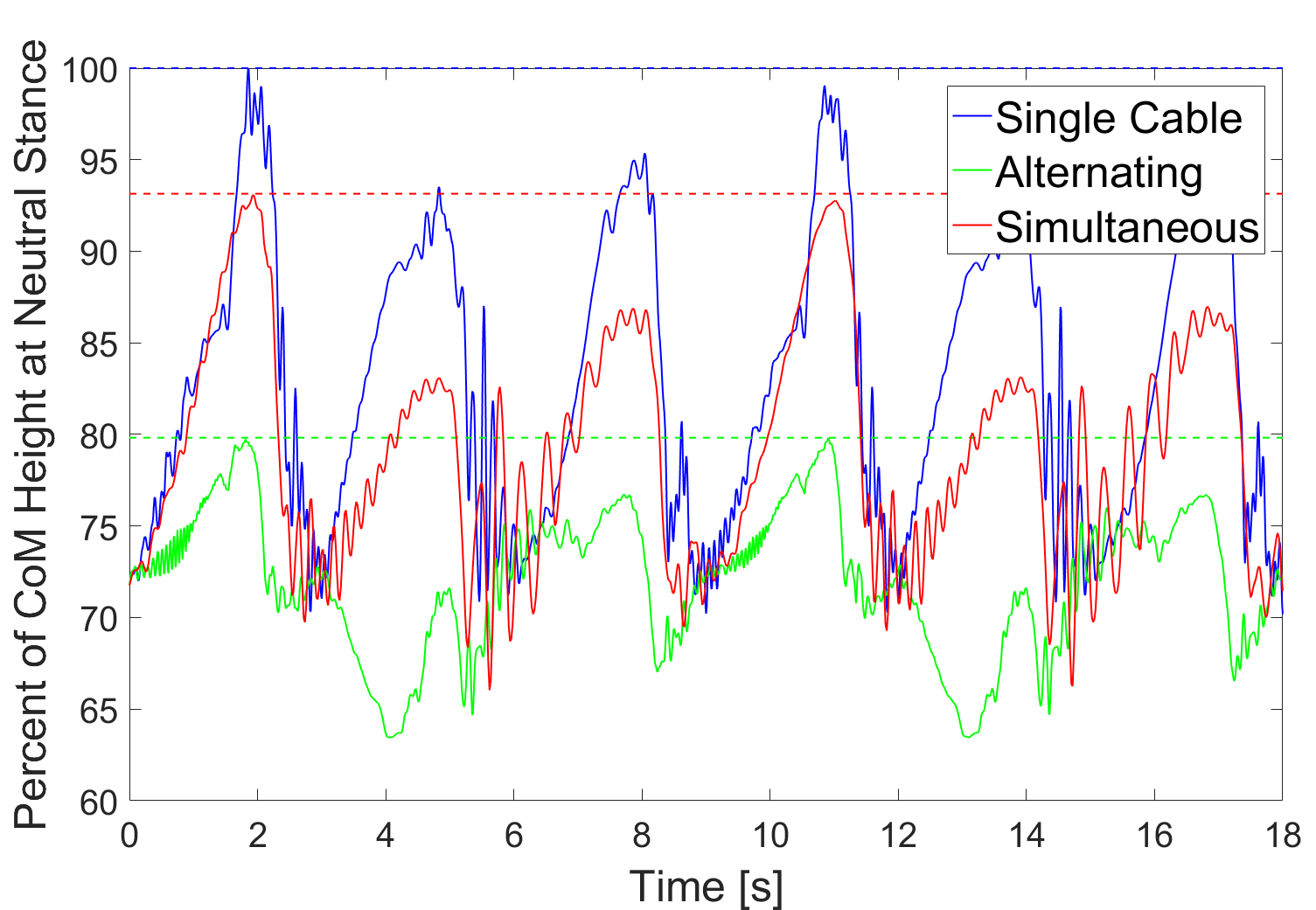}
    \caption{Comparison of robot CoM height over time as a percent of neutral stance CoM height for single-cable and two-cable actuation policies. Maximum heights for each policy shown as dotted lines.}
    \label{fig:com_compare}
    \vspace{-0.4cm}
\end{figure}

\begin{figure}[h]
    \center
    \includegraphics[width=\columnwidth]{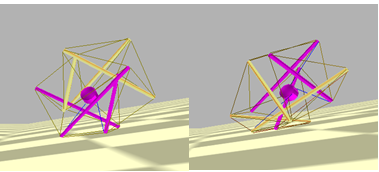}
    \includegraphics[width=\columnwidth]{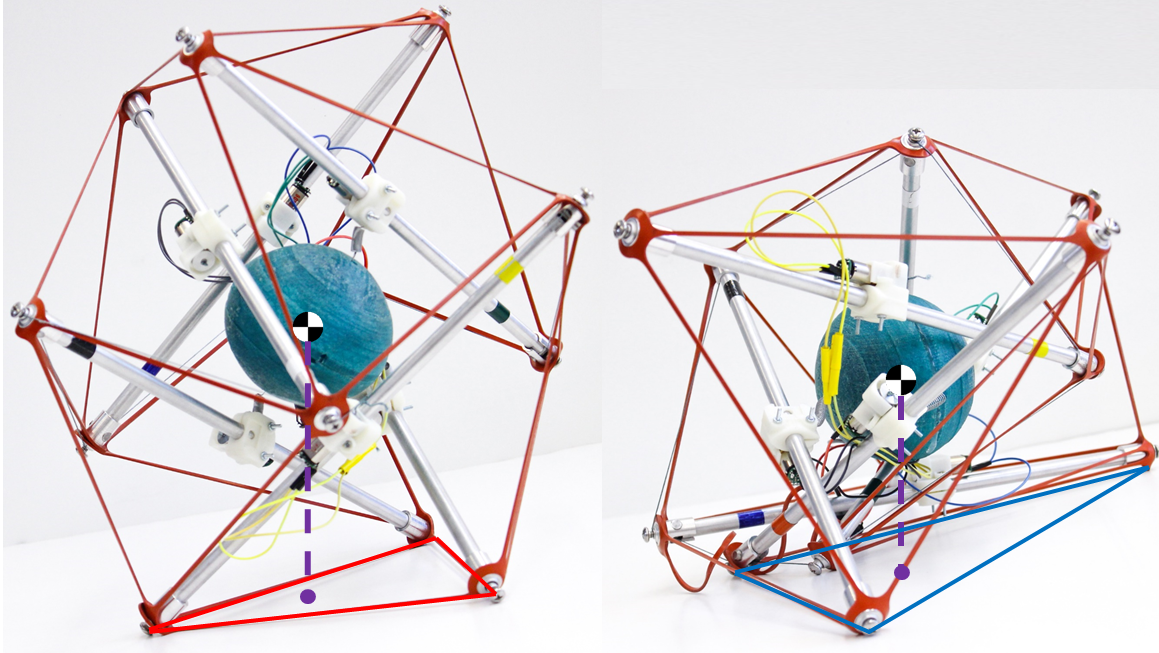}
    \caption{Shown in both simulation and hardware, the payload's CoM height at the robot's neutral state for single-cable actuation (left) is higher than that of the multi-cable actuation policy (right). The base polygon is highlighted in the lower figure.}
    \label{fig:com_compare_photo}
    \vspace{-0.7cm}
\end{figure}

\begin{figure}[tbhp]
    \center
    \includegraphics[width=0.7\columnwidth]{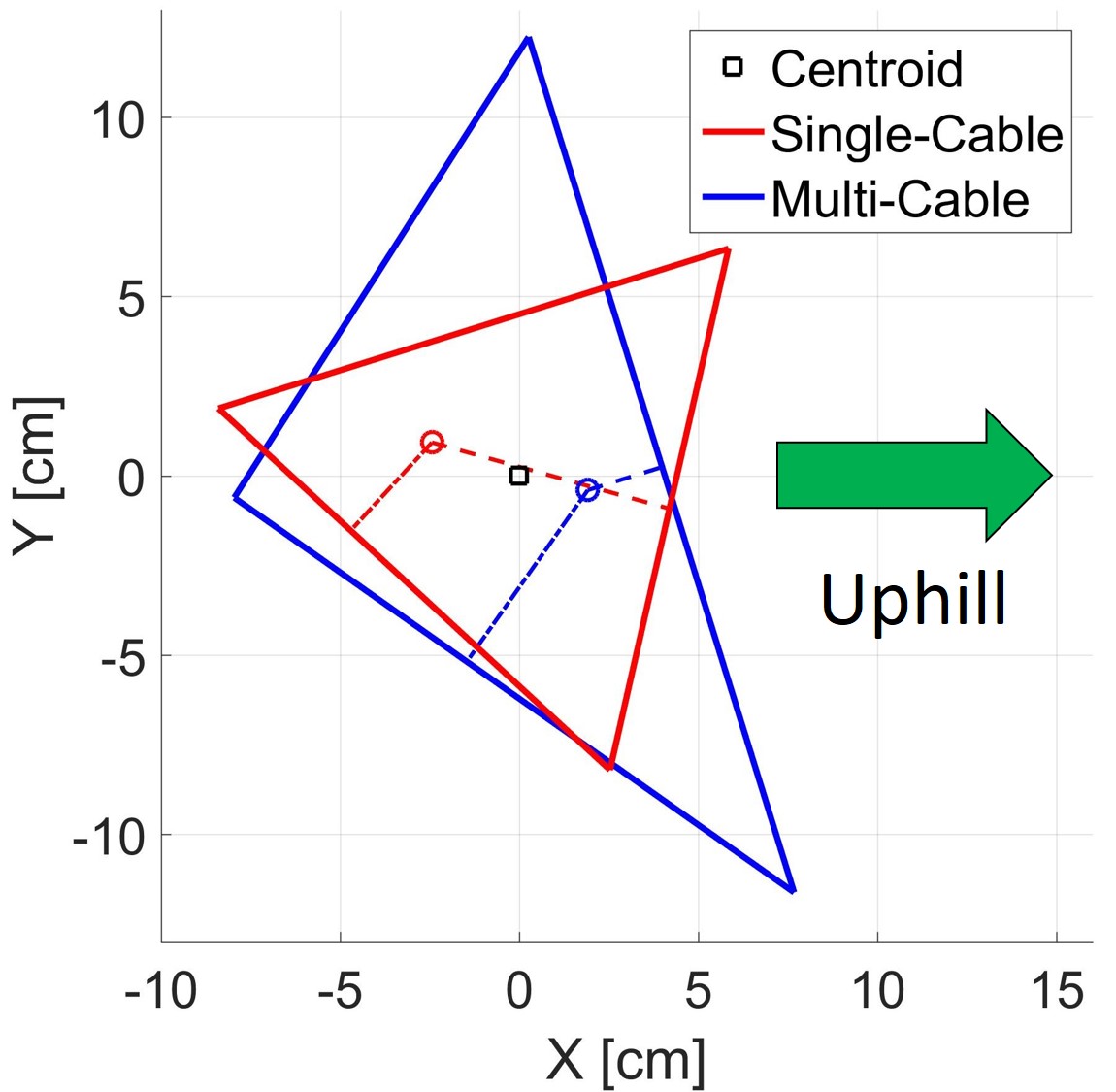} \\
    \caption{Comparison of projected CoM (circles) with supporting base polygons for single-cable (red) and two-cable actuation (blue) policies on a 10\degree~incline. Direction of uphill travel is along the positive $x$ axis. Distance from the uphill edge of the robot's base polygon (dotted lines) is less for multi-cable actuation than single-cable actuation. Similarly, distance from the downhill edge of the robot's base (dash-dotted lines) is greater for the multi-cable policy than the single-cable policy.  }
  \label{fig:BasePolygons}
  \vspace{-0.5cm}
\end{figure}

The significant performance improvements achieved with the two-cable policies are primarily due to the increased stability of the robot and its subsequent ability to avoid rolling downhill during actuation. We believe that this is due to a combination of two primary factors, namely CoM height and number of contact points between the robot and the ground. From the simulation results in Fig.~\ref{fig:com_compare}, it was observed that the average CoM was consistently lower throughout the actuation sequence of the robot, especially at the critical moments approaching the tipping point. On a flat surface, it was found that the maximum CoM heights were 93.1\% and 79.8\% of the neutral stance CoM height for simultaneous, and alternating actuation, respectively. In addition to the lower CoM, both two-cable policies maintain at least one cable in contraction at all times. In contrast to the three contact points in single-cable actuation, the contracted cable keeps the robot in a perpetually forward-leaning stance with four points of contact with the ground, resulting in a larger supporting base polygon (the convex hull of the four contact points), as illustrated by Fig.~\ref{fig:BasePolygons}. Moreover, the stance of robot places the projected CoM uphill of the centroid of the base polygon and farther away from the downhill edge, as opposed to behind it as in the single cable case. This leads to a drastic improvement in incline stability, as the robot is less likely to roll backwards due to external disturbances. Conversely, this also means that it is easier for the robot to roll forwards, as the distance to move the projected CoM outside the supporting polygon in the desired direction is smaller and therefore easier to achieve. This is especially apparent in Fig.~\ref{fig:BasePolygons}, where the CoM is 51.4\% closer to the uphill edge when compared to the single-cable case. The stances of single-cable and two-cable actuation are shown in Fig.~\ref{fig:com_compare_photo}. 

As the robot no longer returns to a neutral state before initiating the next roll sequence, the simultaneous policy saw a notable increase in average speed. However, it did not appear that the increased speed has much effect on the robot's ability to navigate an incline, as the punctuated manner in which actuation is performed means that little if any momentum is preserved from one roll to the next.

As the software incline limits of 24\degree and 26\degree were reached for alternating and simultaneous two-cable actuation respectively, it was found that the robot could no longer reliably navigate the inclined surface, primarily due to insufficient friction. This result was corroborated by our physical hardware experiments.

\subsection{Hardware Experiments of Alternating and Simultaneous Two-Cable Actuation Policies}
In accordance with simulation results, the ability of the robot to actuate multiple cables simultaneously and in alternating order resulted in significant improvements in its ability to navigate steep inclines and achieve high speeds.

The \texorpdfstring{TT-4\textsubscript{mini}}{} was able to leverage alternating two-cable actuation to reliably climb a 24\degree~(44.5\% grade) incline, far outperforming the robot's previous best of 13\degree~(23.1\% grade) set via single-cable actuation. Such a significant improvement establishes this performance as the steepest incline successfully navigated by a spherical tensegrity robot to date. Indeed, the primary cause for failure of two-cable alternating actuation at and beyond 24\degree~was not falling backwards, but rather slipping down the slope due to insufficient friction, in accordance with our measurements mapping the robot's mean coefficient of friction to a theoretical max incline of 26\degree. This suggests that further improvements may be made to the robot's incline rolling ability given careful consideration of material choices in the next design iteration.
 
 \begin{figure}[h]
    \center
    \includegraphics[width=\columnwidth]{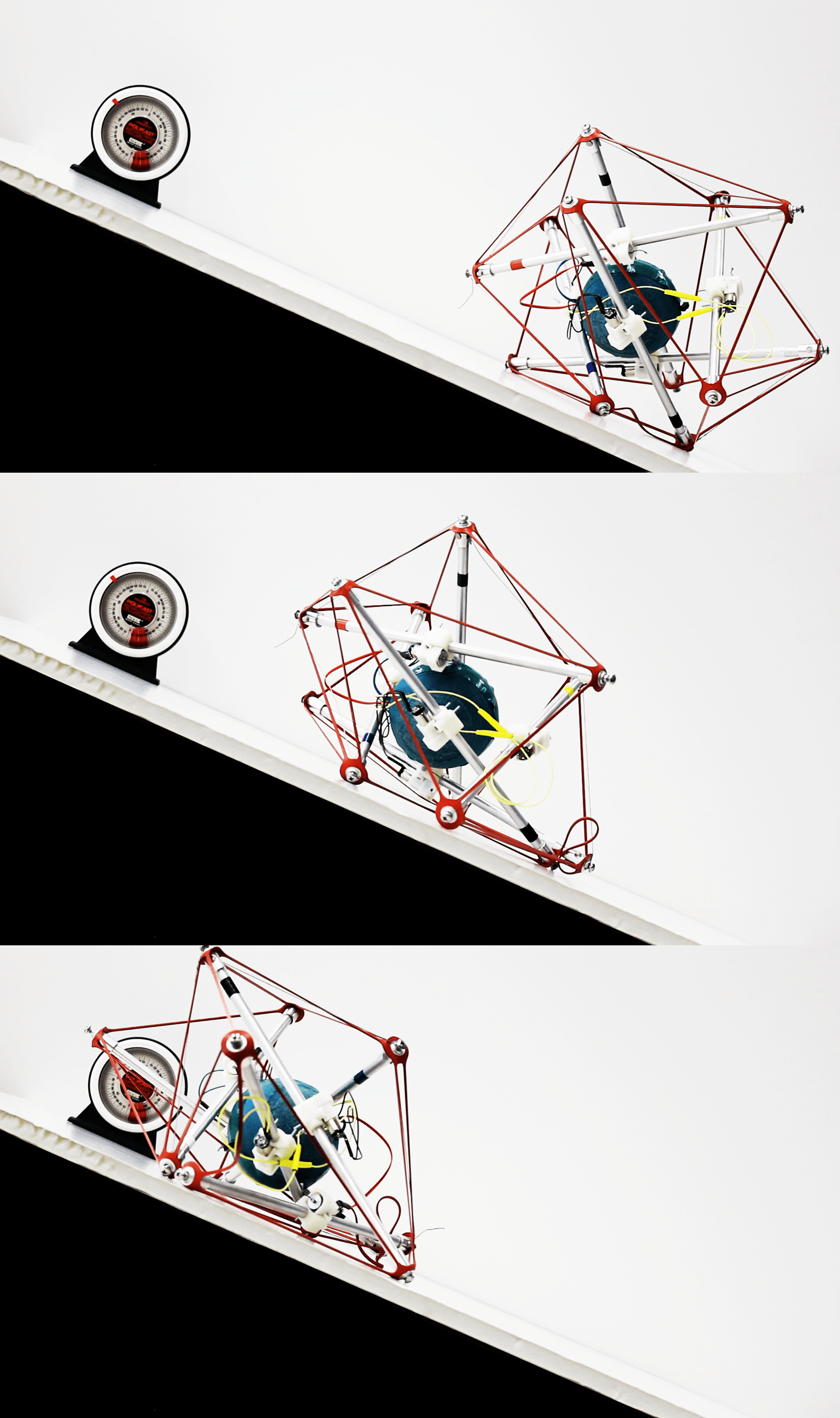}
    \caption{The \texorpdfstring{TT-4\textsubscript{mini}}{} prototype climbing up a 24\degree~incline surface with two-cable alternating actuation.}
    \label{fig:TT4miniIncline}
    \vspace{-0.5cm}
\end{figure}

Not only did the robot's incline climbing performance improve, but its locomotion speed did as well. As mentioned previously, on an incline of 10\degree, the traditional, single-cable actuation policy traveled a distance of 91.4 cm with an average velocity of 1.96 cm/s. However, when performing simultaneous two-cable actuation, the robot was able to travel the same distance with a 10-trial average velocity of 4.22 cm/s, achieving an increase of over 115\% beyond the single-cable baseline. We anticipate that this improvement can be increased by further overlapping the contractions and relaxations of more cables in the simultaneous actuation policy. As the number of cables being simultaneously actuated increases, the rolling pattern increasingly resembles a fluid, spherical roll. However, more complex actuation patterns also require an increasingly skilled robot tele-operator. We recognized that an increase in operator skill leads to an increase in performance, but this also indicates the great potential for intelligent policy optimization and automation. This has the potential to far outperform human operators and achieve ever faster locomotion and the conquering of steeper inclines.

\section{Conclusion}

In this work, we have demonstrated, through both simulation and hardware results, the ability of a spherical tensegrity robot to perform consistent uphill locomotion on steep inclines. This was made possible through the development of a novel multi-cable actuation scheme, which allow the \texorpdfstring{TT-4\textsubscript{mini}}{} to reliably perform forward locomotion on much steeper inclines and at greater speeds than what was previously possible by using only single-cable actuation.

Due to the inherent coupled, nonlinear dynamics of the robot, multi-cable actuation policies render robotic control a challenging intellectual task, providing a launch point for future work. We look forward to exploring the integration of artificial intelligence (particularly evolutionary algorithms and deep reinforcement learning architectures) in this robotic platform to optimize locomotive gaits on varied inclines, and even generate optimal tensegrity topologies, areas which have proven promising in prior work \cite{deeprlnasa,learnedtopology}. We hope to leverage learning algorithms to achieve more fluid and efficient locomotion using a robust and fully autonomous control policy.

\section*{Acknowledgement}

The authors are grateful for funding support from NASA's Early Stage Innovation grant NNX15AD74G.

We also wish to acknowledge the work of the other students on this project: Faraz Ghahani, Carielle U. Spangenberg, Abhishyant Khare, Grant Emmendorfer, Cameron A. Bauer, Kit Y. Mak, Kyle G. Archer, Lucy Kang, Amir J. Safavi, Yuen W. Chau, Reneir Viray, Eirren Viray, and Sebastian Anwar. 

\bibliographystyle{myIEEEtran}
\balance
\bibliography{TensegrityMini.bbl}
\end{document}